\documentclass[3p,times]{elsarticle}

\usepackage{lineno,hyperref}
\usepackage{amsmath}
\usepackage{algorithm}
\usepackage{algorithmicx}
\usepackage{algpseudocode}
\usepackage{graphicx}
\usepackage{multirow}
\usepackage{booktabs}
\usepackage{makecell}
\usepackage{tabularx}
\allowdisplaybreaks[4]
\modulolinenumbers[5]
\usepackage{url}
\journal{Null}
\date{V1.1 - January 8, 2018; V1.3 - December 22, 2018}

\usepackage[figuresright]{rotating}

\begin{document}

\begin{frontmatter}




\title{A generalized concept-cognitive learning: A machine learning viewpoint}


\author[a,b,d]{Yunlong Mi}
\author[b,c,d]{Yong Shi}
\author[e]{Jinhai Li}

\address[a]{School of Computer and Control Engineering, University of Chinese Academy of Sciences, Beijing 101408, China}
\address[b]{Key Laboratory of Big Data Mining and Knowledge Management, Chinese Academy of Sciences, Beijing 100190, China}
\address[c]{Research Center on Fictitious Economy and Data Science, Chinese Academy of Sciences, Beijing 100190, China}
\address[d]{College of Information Science and Technology, University of Nebraska at Omaha, NE 68182, USA}
\address[e]{Faculty of Science, Kunming University of Science and Technology, Kunming 650500, China}

\begin{abstract}
Concept-cognitive learning (CCL) is a hot topic in recent years, and it has attracted much attention from the communities of
formal concept analysis, granular computing and cognitive computing. However, the relationship among cognitive computing (CC), concept-cognitive computing (CCC), CCL and concept-cognitive learning model (CCLM) is not clearly described. To this end, we first explain the relationship of CC, CCC, CCL and CCLM. Then, we propose a generalized concept-cognitive learning (GCCL) from the point of view of machine learning. Finally, experiments on some data sets are conducted to verify the feasibility of concept formation and concept-cognitive process of GCCL.
\end{abstract}


\end{frontmatter}

\correspondingauthor[*]{Corresponding author: YunlongMi. Tel.: +0-000-000-0000 ; fax: +0-000-000-0000 .}
\email{YunlongMi@yeah.net}


\newtheorem{definition}{Definition}
\newtheorem{theorem}{Theorem}
\newtheorem{proposition}{Proposition}
\newtheorem{property}{Property}
\newtheorem{proof}{Proof}

\section{Introduction}
Cognitive computing (CC) is known as an artificial intelligence of computer system modeled on the human brain \cite{Modha2011Cognitive,Wang2009On}. It has been researched by different aspects, such as memory \cite{Li2015Spontaneous,Srivastava2017A}, learning \cite{Feldman2000Minimization,Lake2015Human,Li2015Concept,Zhao2016Cognitive}, language \cite{Peng2017Incrementally}, thinking \cite{Wang2006Cognitive}, objective \cite{Cox2017Goal,Grace2016Surprise} and problem solving \cite{Newell1972Human,Wang2010On,Zadeh1979Fuzzy}.

A \emph{concept} is a cognitive unit to identify a real-world concrete entity or model a perceived-world abstract subject by its extent and intent \cite{Wang2008On,Wang2009On}. Up to now, for meeting different requirements of data mining, various concepts have been proposed such as abstract concepts \cite{Wang2008On}, Wille¡¯s concepts \cite{Wille1982Restructuring}, property-oriented concepts \cite{D¨¹ntsch2002Modal}, object-oriented concepts \cite{Ma2017Concept,Yao2004A,Yao2004Concept}, AFS-concepts \cite{Wang2008Concept}, approximate concepts \cite{Li2015Concept} and three-way concepts \cite{Li2017Three}. Concept learning is to learn unknown concepts by a certain approach from a given clue such as concept algebra system, queries, cognitive system, cloud model, set approximation, iteration, etc \cite{Li2017Three}. From the viewpoint of Yao \cite{Yao2009Interpreting}, concept learning can be understood from three aspects: the abstract level, brain level and machine level. Moreover, granular computing \cite{Pedrycz2013Granular,Qian2010MGRS,Zadeh1979Fuzzy} has been combined with it to deal with complex problems \cite{Kang2012Formal,Kumar2015Formal,Ma2007Granular,Ma2015Rough,Wu2009Granular}.

\emph{Concept-cognitive learning (CCL)} is to learn and apply concepts via its extent and intent by simulating human brain. It was firstly researched from an abstract perspective by Zhang and Xu \cite{Zhang2007Cognitive}, Wang \cite{Wang2008On} and Yao \cite{Yao2009Interpreting}. Then, it was investigated by different aspects to meet different requirements in the real world \cite{Li2015Cognitive,Li2015Concept,Mi2018Research,Niu2017Parallel,Xu2014A,Xu2016Granular}. Although many attempts are made to study CCL, there is still lack of a generalized framework for CCL.  The main contributions of this paper are to (1) expound the relationship among CC, concept-cognitive computing (CCC), CCL and concept-cognitive learning model (CCLM), and (2) explore CCL from viewpoint of machine learning and then propose a generalized concept-cognitive learning (GCCL).

The remainder of this paper is organized as follows. Section 2 briefly recalls some basic concepts related to GCCL; moreover, the notion of CCC is proposed. Section 3 establishes a framework of CCL from the point of view of machine learning.  Section 4 conducts some experiments to assess concept formation and concept-cognitive process of the propsoed GCCL. The paper is then concluded in section 5 with a brief summary.

\section{Preliminary Knowledge}
In this section, we first show the relationship among CC, CCC, CCL and CCLM, and then some related concepts of GCCL are briefly described.
\subsection{Concept-cognitive Computing}
\begin{definition}[Cognitive Computing \cite{Modha2011Cognitive,Wang2009On}]
Cognitive computing (CC) is an emerging computing paradigm of intelligent science that implements computational intelligence, which to achieve different levels of perception, memory, learning, language, thinking, objective, problem solving, etc., by trying to solve the problems of imprecision, uncertainty and partial truth in biological system.
\end{definition}

\begin{definition}[Concept-cognitive Computing]
Concept-cognitive computing (CCC) is a form of cognitive computing that mainly includes a knowledge
storage mechanism (KSM), a learning mechanism (LM), a goal operation mechanism (GOM), a problem solving
mechanism (PSM), a generative mechanism (GM), etc., where the KSM is based on the concept lattice space
which consists of various concepts conforming to a Galois connection and the LM can dynamically learn
some new concepts (or knowledge) from the extent and intent of concepts in complex environments.
\end{definition}

Based on the KSM and LM, we believe that the CCC can achieve computational intelligence such as memory, thinking, objective and problem solving.

In Definition 2, the CCC takes concepts as the knowledge carrier and then achieves some cognitive goals such as memory
and learning. Note that, here, we focus on knowledge storage and learning because, to some extent, other
cognitive goals in CCC are influenced by them. Therefore, from the perspectives of cognitive computing and machine learning, we make
further research on the GCCL that only consists of the KSM and LM. The relationship of CC, CCC, CCL and CCLM is described as Fig. 1.
\begin{figure}
\setlength{\abovecaptionskip}{0.cm}
\setlength{\belowcaptionskip}{-0.cm}
\centering
\includegraphics[width=0.5\textwidth]{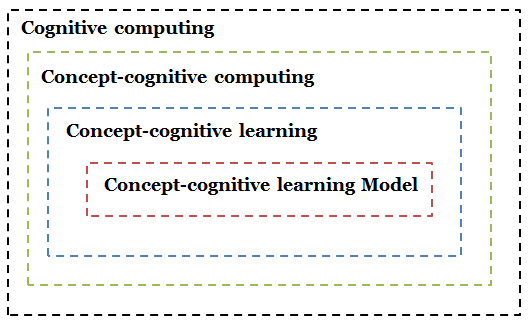}
\caption{Relationship among CC, CCC, CCL and CCLM. CCC contains CCL, and is included in CC.}
\label{Fig1:relationship}
\end{figure}

Here, it is worth stressing that the notion of \emph{concept-cognitive learning}. we know that CCL was studied from a cognitive perspective by many scholars \cite{Kumar2015Formal,Li2015Cognitive,Li2015Concept,Li2017Three,Mi2018Research,Niu2017Parallel,Zhang2007Cognitive,Zhao2016Cognitive}, and to meet of different requirements, various of names were used such as cognitive model \cite{Zhang2007Cognitive}, cognitive system \cite{Xu2014A} and cognitive concept learning \cite{Li2015Cognitive,Zhao2016Cognitive}; however, in order to stress the importance of the \emph{concept} in machine learning and cognitive process, its name was also discussed by Mi and Li in 2016; and then we first used the concept of \emph{concept-cognitive learning} in 2016 \cite{Mi2018Research} and it is called classical CCL; moreover, an atomic concept and a composite concept were also defined in \cite{Mi2018Research}. Based on it, inspired by the similarity degree \cite{Yu:Machine} and a formal decision context \cite{Zhang2005Uncertain}, Mi et. al began to explore how to incorporate it into machine learning in 2016, and then a novel CCLM was proposed in 2018 \cite{Shi-Mi-Li-Liu:Concept}. In this case, GCCL can be considered to consist of two parts: 1) classical CCL, and 2) CCLM, in which different real tasks such as classification task, knowledge reasoning, and concept generation will be achieved. Hereafter, we call GCCL as CCL when there is no confusion.

In addition, we show two main features of classical CCL as follows:
\begin{itemize}
  \item the concepts being employed as information carries in CCL, in which concepts consist of some real concepts or many virtual concepts;
  \item the cognitive process in CCL being completed in a dynamic environment (or by dynamic learning) instead of learning completed by only once (or static learning).
\end{itemize}
And the main features of CCLM include the two features of classical CCL and the concept similarity (CS) degree among concepts.
\subsection{Concept-cognitive Learning for Learning}
To facilitate the understand of CCL, this subsection briefly reviews some related concepts. Although there is still no an accepted unified theory for CCL, we believe that the approximate CCL is a classical CCL system. Thus, we show that the classical cognitive operator and concept-cognitive process of approximate CCL as follows. 
\begin{definition}[Classical Cognitive Operator {\cite{Li2015Concept}}]
 Let $G$ be an object set and $M$ be an attribute set. We denote the power sets of $G$ and $M$ by $2^{G}$ and $2^{M}$,
 respectively. The classical cognitive operators are two set-valued mappings:
\begin{equation}
\begin{gathered}
\mathcal{F}:2^{G}\rightarrow 2^{M},\\
\mathcal{H}:2^{M}\rightarrow 2^{G}
\end{gathered}
\end{equation}
where $\mathcal{F}$ denotes object-attribute operator and $\mathcal{H}$ denotes attribute-object operator. If for any
$A_{1},A_{2}\subseteq G$ and $B\subseteq M$, the following properties hold:
\begin{align*}
(\romannumeral1) &&&A_{1}\subseteq A_{2}\Rightarrow \mathcal{F}(A_{2})\subseteq \mathcal{F}(A_{1}),\\
(\romannumeral2) &&&\mathcal{F}(A_{1}\cup A_{2})\supseteq \mathcal{F}(A_{1})\cap \mathcal{F}(A_{2}),\\
(\romannumeral3) &&&\mathcal{H}(B)=\{g\in G|B\subseteq \mathcal{F}(\{g\})\}.
\end{align*}
\indent Here, $\mathcal{F}(\{g\})$ are rewritten as $\mathcal{F}(g)$ for short when there is no confusion.
\end{definition}

\begin{definition}[Cognitive Concept \cite{Li2015Concept}]
Let $\mathcal{F}$ and $\mathcal{H}$ be cognitive operators. For $g\in G$ and $m\in M$, we say that $(\mathcal{H}\mathcal{F}(g),
\mathcal{F}(g))$ and $(\mathcal{H}(m),\mathcal{F}\mathcal{H}(m))$ are granular concepts. Then, any cognitive concept can be
obtained by granular concepts, where for $A\subseteq G$ and $B\subseteq M$, if $\mathcal{F}(A)=B$ and $\mathcal{H}(B)=A$,
the ordered pair $(A,B)$ is called a cognitive concept.
\end{definition}

In what follows, note that the set of all cognitive concepts is called a \emph{concept space}.

\begin{definition}[Concept-cognitive Process \cite{Li2015Concept}]
Let $G_{i-1},G_{i}$ be object sets of $\{G_{t}\}\uparrow$, where $\{G_{t}\}\uparrow$ is a non-decreasing sequence of object sets $G_{1},G_{2},...,G_{n}$
, and $M_{i-1},M_{i}$ be attribute sets of $\{M_{t}\}\uparrow$, where $\{M_{t}\}\uparrow$ is a non-decreasing sequence of attribute sets $M_{1},M_{2},...,M_{m}$.
Denote $\Delta G_{i-1}=G_{i}-G_{i-1}$ and $\Delta M_{i-1}=M_{i}-M_{i-1}$. Suppose
\begin{align*}
(\romannumeral1) &&&\mathcal{F}_{i-1}:2^{G_{i-1}}\rightarrow 2^{M_{i-1}},&& \mathcal{H}_{i-1}:2^{M_{i-1}}\rightarrow 2^{G_{i-1}},\\
(\romannumeral2) &&&\mathcal{F}_{\Delta G_{i-1}}:2^{\Delta G_{i-1}}\rightarrow 2^{M_{i-1}}, &&\mathcal{H}_{\Delta G_{i-1}}:2^{M_{i-1}}\rightarrow 2^{\Delta G_{i-1}},\\
(\romannumeral3) &&&\mathcal{F}_{\Delta M_{i-1}}:2^{G_{i}}\rightarrow 2^{\Delta M_{i-1}},&&\mathcal{H}_{\Delta M_{i-1}}:2^{\Delta M_{i-1}}\rightarrow 2^{G_{i}},\\
(\romannumeral4) &&&\mathcal{F}_{i}:2^{G_{i}}\rightarrow 2^{M_{i}},&&\mathcal{H}_{i}:2^{M_{i}}\rightarrow 2^{G_{i}}
\end{align*}
are four pairs of cognitive operators satisfying the following properties:
\begin{align}
\mathcal{F}_{i}(g)&=
    \begin{cases}
    \mathcal{F}_{i-1}(g)\cup \mathcal{F}_{\Delta M_{i-1}}(g), &\text{if}\ g\in G_{i-1},\\
    \mathcal{F}_{\Delta G_{i-1}}(g)\cup \mathcal{F}_{\Delta M_{i-1}}(g), &\text{otherwise},
    \end{cases}\\
\mathcal{F}_{i}(m)&=
    \begin{cases}
    \mathcal{H}_{i-1}(m)\cup \mathcal{H}_{\Delta G_{i-1}}(m), &\ \text{if}\ m\in M_{i-1},\\
    \mathcal{H}_{\Delta M_{i-1}}(m), &\ \text{otherwise},
    \end{cases}
\end{align}
where $\mathcal{F}_{\Delta G_{i-1}}(g)$ and $\mathcal{H}_{\Delta G_{i-1}}(m)$ are set to be empty when $\Delta G_{i-1}=\emptyset$, and $\mathcal{F}_{\Delta M_{i-1}}(g)$ and $\mathcal{H}_{\Delta M_{i-1}}(m)$ are set to be empty when $\Delta M_{i-1}=\emptyset$. Then we say that $\mathcal{F}_{i}$ and $\mathcal{H}_{i}$ are extended cognitive operators of $\mathcal{F}_{i-1}$ and $\mathcal{H}_{i-1}$ with the newly input information $\mathcal{F}_{\Delta G_{i-1}}$,$\mathcal{H}_{\Delta G_{i-1}}$ and $\mathcal{F}_{\Delta M_{i-1}}$,$\mathcal{H}_{\Delta M_{i-1}}$.
\end{definition}

Here, The cognitive operators are given in Definition 3, and the concept of granular concept is described in Definition 4. Based on Definitions 3 and 4, the basic mechanism of concept-cognitive process is shown in Definition 5.

\section{Framework of Concept-cognitive Learning}
In this section, we describe the CCL from the perspective of machine learning, which includes a framework of generalized CCL and a concrete example (i.e., approximate CCL).
\subsection{Generalized Concept-cognitive Learning}
A GCCL can be described as four aspects: 1) concept formation, 2) concept storage, 3) concept-cognitive process and 4) concept application. The phase of concept mapping in Fig. 2 includes concept constructing and mapping it into different sub-concept spaces, and then it takes the concept space as knowledge storage carrier. When facing some newly input data, according to Definition 5, the current concept space will be updated with newly formed concept. Based on the formed concept space, we can do some meaningful concept applications such as learning the lower and upper approximation from a given concept, concept classification, and rules extraction. Note that the concept-cognitive process in Fig. 2 is a recursive process in GCCL.

Simultaneously, we believe that the GCCL consists of the classical CCL such as approximate CCL \cite{Li2015Concept}, three-way CCL \cite{Li2017Three} and parallel CCL \cite{Mi2018Research,Niu2017Parallel}, and different CCLMs based on a regular formal decision context \cite{Shi-Mi-Li-Liu:Concept}. For instance, there is a CCLM for incremental learning \cite{Shi-Mi-Li-Liu:Concept}, a concurrent CCLM that is based on multi-threads, a semi-supervised CCLM for semi-supervised learning and so on. Now, we take the approximate CCL for example as follows.
\begin{figure}
\setlength{\abovecaptionskip}{0.cm}
\setlength{\belowcaptionskip}{-0.cm}
\centering
\includegraphics[width=0.93\textwidth]{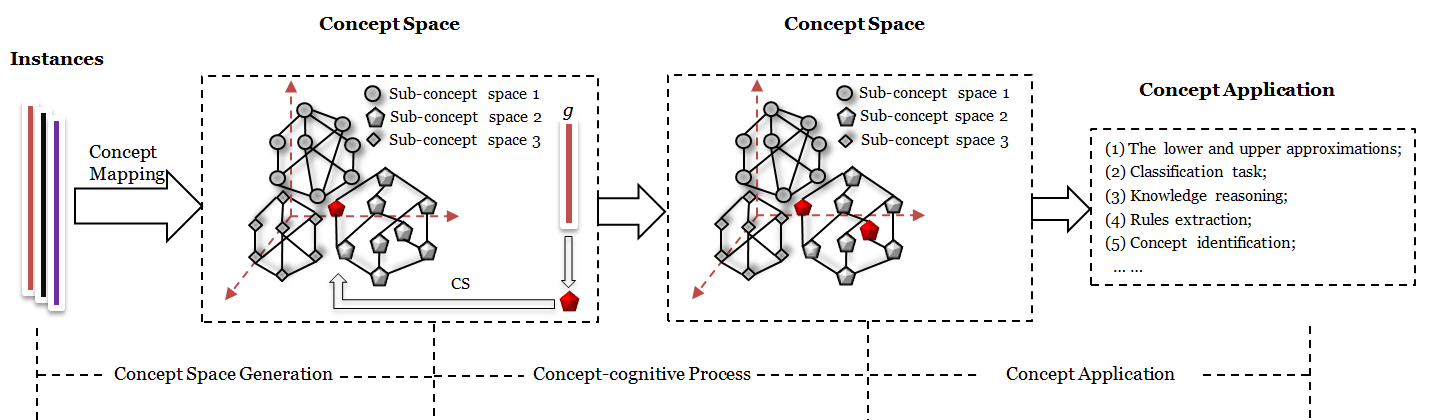}
\caption{Framework of GCCL includes three phases: 1) Initial concept space generation;  2) Concept-cognitive process; 3) Concept application. The CS denotes concept similarity degree.}
\label{Fig2:generalized CCL}
\end{figure}

\subsection{Approximate Concept-cognitive Learning}
Since the cognitive concept learning proposed by Li et al. \cite{Li2015Cognitive,Li2015Concept,Zhao2016Cognitive} mainly focuses on the lower and upper approximation from a given concept, it is called as an approximate concept-cognitive learning (CCL). Fig. 3 provides a framework of approximate CCL based on a formal context. As a concrete of GCCL, the main features of approximate CCL perform two aspects:
\begin{itemize}
  \item The results of approximate CCL are to learn one exact or two approximate cognitive concepts from a given object set, an attribute set or a pair of object and attribute sets.
  \item Different from CCLM, all data are only mapped into one concept lattice space in approximate CCL; in other words, data will be mapped into different sub-concept spaces in CCLM.
\end{itemize}

\begin{figure}
\centering
\includegraphics[width=0.60\textwidth]{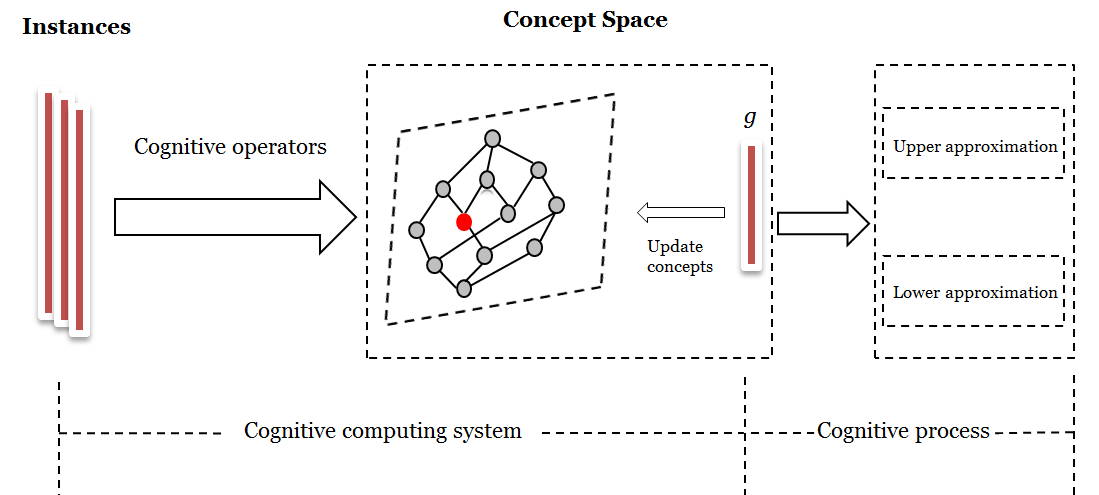}
\caption{Framework of an approximate concept-cognitive learning.}
\label{Fig3:an approximate concept-cognitive learning}
\end{figure}

\section{Experimental Evaluation of Generalized Concept-cognitive Learning}
In this section, we conduct some experiments to evaluate the GCCL that only involves two aspects: concept formation and concept-cognitive process. Our main aim is to verify the feasibility of concept space generation and concept-cognitive process in GCCL; therefore, there are only some simple experiments.  We implement GCCL process in the following settings:  AMD FX(tm)-4300 Quad-Core Processor 3.80 GHz CPU, 4GB main memory, JDK: jdk1.8.0\_20, Eclipse: eclipse-4.2.
And we code our algorithms in Java.

In the experiments, two data sets are selected from UCI Machine Learning Repository \cite{Frank2010UCI} to achieve the evaluation task; namely, Voting Records
data set and Mushroom data set. Since the two data sets can not directly be used to evaluate the GCCL, the data pre-processing technique and nominal scale are required to convert them. In what follows, we introduce the detailed usage of two data sets.

(1) Voting Records data set has 435 instances and 16 attributes. It is converted by the scaling approach \cite{Ganter1999Formal} into a standard data set. Then, 20, 50, and 100 instances are respectively used as concept formation. For convenience, we denote them by Data sets 1, 2, and 3, respectively.

(2) Mushroom data set consists of 8,124 instances and 22 attributes. The scaling approach is applied to the data set for generating a standard one. Then, we take 200, 500, 1000, and 2000 instances as concept formation, respectively. For brevity, we denote them by Data sets 4, 5, 6, and 7.

Here, we only consider that the information on the object set $G$ will be updated as time goes by. Namely, $\mathcal{F}_{\Delta M_{i-1}}(g)$ and $\mathcal{H}_{\Delta M_{i-1}}(m)$ are empty. For brevity, we write original attributes and new attributes by scaling approach as attributes (o) and attributes (s), respectively. See Table 1 for details. In Table 2, the item \#10 instances means conducting 10 instances for incremental learning based on initial concepts, and it has the same meaning for item \#100 instances (or \#1000 instances). The results for concept lattice size and running time (including initial concepts and incremental learning) are reported in Table 1 and Table 2, respectively. Finally, it can be observed from the Tables 1 and 2 that the GCCL (only considering two parts: concept formation and concept-cognitive process) are feasible for the seven chosen data sets.
\begin{table}[h]
\footnotesize
\caption{Size of concept lattice for GCCL.}
\begin{tabular*}{\hsize}{@{\extracolsep{\fill}}lllll@{}}
\hline
Data set & \#initial instances & \#attributes (o) &  \#attributes (s) & \#the size of concept lattice space \\
\hline
 Data set 1 &\qquad 20   &\qquad 16 &\qquad 32  &\qquad 55   \\
 Data set 2 &\qquad 50   &\qquad 16 &\qquad 32  &\qquad 97   \\
 Data set 3 &\qquad 100  &\qquad 16 &\qquad 32  &\qquad 144  \\
 Data set 4 &\qquad 200  &\qquad 22 &\qquad 128 &\qquad 311  \\
 Data set 5 &\qquad 500  &\qquad 22 &\qquad 128 &\qquad 628  \\
 Data set 6 &\qquad 1,000 &\qquad 22 &\qquad 128 &\qquad 1,141 \\
 Data set 7 &\qquad 2,000 &\qquad 22 &\qquad 128 &\qquad 2,149 \\
\hline
\end{tabular*}
\label{Tab1:CCL-size}
\end{table}

\begin{table}[h]
\footnotesize
\caption{Running time (seconds) of incremental learning for GCCL.}
\begin{tabular*}{\hsize}{@{\extracolsep{\fill}}lllll@{}}
\hline
Data set & \#initial time & \#10 instances &  \#100 instances & \#1000 instances \\
\hline
 Data set 1 &\qquad 0.0659  &\qquad 0.2186 &\qquad 0.3658 &\qquad --\\
 Data set 2 &\qquad 0.1122  &\qquad 0.2177 &\qquad 0.3734 &\qquad --\\
 Data set 3 &\qquad 0.2248  &\qquad 0.2192 &\qquad 0.4070 &\qquad --\\
 Data set 4 &\qquad 2.7160  &\qquad 0.2576 &\qquad 0.8398 &\qquad 12.8152\\
 Data set 5 &\qquad 9.3878  &\qquad 0.2800 &\qquad 1.1008 &\qquad 15.1078\\
 Data set 6 &\qquad 26.0752 &\qquad 0.3162 &\qquad 1.5070 &\qquad 20.0294\\
 Data set 7 &\qquad 81.2578 &\qquad 0.4056 &\qquad 2.4696 &\qquad 28.6650\\
\hline
\end{tabular*}
\label{Tab1:CCL-time}
\end{table}

\section{Conclusion}
In this paper, we firstly give a notion of CCC and elucidate the relationship among CC, CCC, CCL and CCLM.  Whereafter, a framework for GCCL is proposed. Finally, some experiments on real world data have proved the feasibility of concept generation and concept-cognitive process in GCCL.
Note that to stress the importance of concept carries in cognitive learning and integrate it with meachine learning, the notion of \emph{concept-cognitive learning} is used.

Based on it, more details for the GCCL with different concept applications (for example, classification, regression, concept identification and others) will be provided in the further research tasks.











\bibliographystyle{plain}
\bibliography{gccl}

\end{document}